\documentclass[sigconf]{acmart}

\usepackage{subcaption}
\usepackage{amsmath}

\AtBeginDocument{%
  \providecommand\BibTeX{{%
    \normalfont B\kern-0.5em{\scshape i\kern-0.25em b}\kern-0.8em\TeX}}}

\begin{document}

\title{Leveraging Uncertainty in Deep Learning for Selective Classification}

\author{Mehmet Yigit Yildirim}
\affiliation{%
  \institution{Arizona State University}
  \city{Tempe}
  \state{AZ}
}
\email{yigityildirim@asu.edu}

\author{Mert Ozer}
\affiliation{%
  \institution{Arizona State University}
  \city{Tempe}
  \state{AZ}}
\email{mozer@asu.edu}

\author{Hasan Davulcu}
\affiliation{%
  \institution{Arizona State University}
  \city{Tempe}
  \state{AZ}}
\email{hdavulcu@asu.edu}

\renewcommand{\shortauthors}{Yildirim et al.}

\begin{abstract}
The wide and rapid adoption of deep learning by practitioners brought unintended consequences in many situations such as in the infamous case of Google Photos' racist image recognition algorithm; thus, necessitated the utilization of the quantified uncertainty for each prediction. There have been recent efforts towards quantifying uncertainty in conventional deep learning methods (e.g., dropout as Bayesian approximation); however, their optimal use in decision making is often overlooked and understudied. In this study, we propose a mixed-integer programming framework for classification with reject option (also known as \textit{selective classification}), that investigates and combines model uncertainty and predictive mean to identify optimal classification and rejection regions. Our results indicate superior performance of our framework both in non-rejected accuracy and rejection quality on several publicly available datasets. Moreover, we extend our framework to cost-sensitive settings and show that our approach outperforms industry standard methods significantly for online fraud management in real-world settings. 

\end{abstract}



\maketitle

\section{INTRODUCTION}

Machine learning classifiers are far from outputting perfect results due to several reasons: data quality, feature informativeness, model selection, and hyper-parameter tuning are just some of the factors contributing to the variability of the outcomes. Although well-trained models offer a high level of accuracy on the macro level, making confident inferences for individual instances is difficult, nevertheless necessary.

Bayesian literature offers a rich set of classification techniques \cite{bayesiandeep,gp_book} for jointly quantifying uncertainty and prediction at inference level. A recent application of dropout neural networks as Bayesian approximation of deep Gaussian Process by Gal and Ghahramani open a new avenue of quantifying uncertainty in traditional deep learning settings where a simple dropout mechanism is applicable \cite{gal2016dropout}.

The gained ability to effectively represent the uncertainty within existing deep learning architectures has been an essential step for democratizing AI safety \cite{amodei2016concrete}. Nevertheless, the following question remains open: how can one make use of the model uncertainty to make optimal decisions? The approach we focus on in this study is called \textit{selective classification} also known as \textit{classification with reject option} where the classifier rejects making a decision when uncertain.

Selective classification is critical for many applications, and the concept of ``rejection" can have different meanings in various contexts. In medical diagnosis, a doctor might order diagnostic tests before making a decision. In fraud management, an expert human analyst would start a manual investigation. In self-driving cars, the human driver would be given control to operate the vehicle. In all cases, rejecting most of the instances would defeat the purpose, and being inaccurate could result in fatal consequences. Hence, a practical framework for selective classification must be able to operate accurately under defined rejection capacity constraints.

A recent study in the medical domain \cite{leibig2017leveraging} has demonstrated the potential of the model uncertainty for selective classification. However, the authors' utilization of the measure is solely based on a simple ranking of it, which makes their work unsuitable for many online or streaming settings. To the best of our knowledge, how model uncertainty compares to or interacts with the more traditional ways of conducting selective classification such as using Bayes risk \cite{chow1970optimum} has not been explored. 

Hence, we propose a Mixed-Integer Programming (MIP) formulation for selective classification called MIPSC to address these requirements. MIPSC finds optimal classification and rejection regions by investigating the relationship between the model uncertainty and predictive mean with the desired rejection capacity without having to define arbitrary rejection costs. Furthermore, we develop cost-sensitive extensions to our MIP model and exhibit the framework's extensibility and usability in real-world problems such as fraud management, where defining domain-specific and example-dependent costs are necessary.

Main contributions of this paper are:
\begin{enumerate}
    \item Introducing the first mixed integer programming solution for selective classification,
    \item Utilizing predictive mean and model uncertainty of dropout NNs for optimal decision making,
    \item Presenting an online fraud management case in a real-world setting.
\end{enumerate}

Rest of the paper is organized as follows: Section 2 gives a brief background on uncertainty representation in AI, selective classification, and mixed-integer programming. In Section 3, we describe our selective classification model, MIPSC, and describe how to extend it to cost-sensitive settings. Then, we demonstrate the effectiveness of both our cost-insensitive and cost-sensitive models in Section 4. Finally, we conclude our paper in Section 5. 

\section{RELATED WORK}

\subsection{Uncertainty Representation and Applications}

Many practitioners and researchers make use of the probability outputs from the trained model (i.e., softmax output in deep learning) as an uncertainty measure; however, many classifiers output distorted probabilities \cite{niculescu2005obtaining} and may lead to misleading actions. Moreover, even when corrected by proposed probability calibration methods such as Isotonic Regression \cite{kruskal1964nonmetric} or Platt Scaling \cite{platt1999probabilistic}, posterior probabilities as point estimates lack the detail and information to provide a correct interpretation of the model uncertainty. So, Bayesian approaches such as \cite{bayesiandeep,gp_book} are the intuitive methods to quantify and represent the model uncertainty correctly. Due to the computational complexity of the Bayesian methods, Gal and Ghahramani propose using Monte Carlo sampling over dropout neural networks as an approximation to Bayesian inference \cite{gal2016dropout}. This approach's effectiveness is demonstrated in a medical-domain application \cite{leibig2017leveraging}. Our work builds upon this framework by combining model uncertainty and predictive mean optimally for classification with reject option.

\subsection{Selective Classification}

Selective classification or classification with reject option has been studied since the 1970s and it has started gaining traction again in the recent decade. It is defined as giving an option to the classifier to express uncertainty and to reject making a certain prediction. Chow's work \cite{chow1970optimum}, being the first study in the field, introduces the concept and proposes a decision-theoretic framework to find the Bayesian-optimal reject threshold. Tortorella \cite{tortorella2000optimal}, and Santos-Pereira and Peres \cite{SANTOSPEREIRA2005943} propose cost-sensitive learning extensions to classification with reject option methods with arbitrary cost-functions. Herbei and Wegkamp \cite{herbei2006classification} develop excess risk bounds for the classification with a reject option for both cost-sensitive and cost-insensitive cases. On the other hand, El-Yaniv and Wiener \cite{el2010foundations} find these cost models unsuitable as it is difficult to quantify the cost of rejection in many cases. Instead, authors focus on theoretical risk-coverage (RC) trade-off without considering explicit costs. Researchers have been adapting this idea to different classifiers, and recently Geifman and El-Yaniv's work \cite{yanivdeep} on modifying deep neural networks for selective classification was proposed. Our work differs fundamentally from Geifman and El-Yaniv's work \cite{yanivdeep} by (1) not being built-in within the deep neural network itself; so it becomes compatible with any existing trained models and systems, and (2) utilizing dropout MC sampling for uncertainty estimation. 

\subsection{Mixed-Integer Programming}
Mixed-Integer programming (MIP) is a powerful modeling tool that has been around for decades. MIP has been commonly utilized by the operations research community; however, practitioners and researchers from other domains hesitated to adopt it due to its computational and theoretical complexity \cite{bixby2010mixed}. During the last three decades, algorithmic advances in integer optimization combined with hardware improvements have enabled a 200 billion factor speedup in solving MIP problems \cite{bertsimas2016best}. Now, mixed integer linear techniques are viewed as mature, fast, and robust; thus are applied to the problems with up to millions of variables \cite{geissler2012using}. Machine learning community also started employing MIP techniques in several problems, such as for optimal feature selection\cite{bertsimas2016best} and for deriving interpretable machine learning algorithms \cite{goh2014box}. The key factors for our decision to use an MIP formulation are (1) its ability to naturally express the problem, the objective, and the constraints, (2) its capability to provide an exact optimal solution, and (3) its ease of extensibility to more specific settings.

\section{PROPOSED MODELS}
In this work, we propose a mixed integer programming model which finds optimal regions in deep neural network classifier output to reject making a classification. To take not only the output of the deep neural network classifier but also its uncertainty into consideration, we choose to use dropout NNs (DNN) \cite{gal2016uncertainty} throughout our modeling and experiments. Dropout NNs have been proven to approximate deep Gaussian processes which generate predictive mean($\mu$) and model uncertainty($\sigma$) in the form of standard deviation. In the following sections, we explain how we make use of both outputs (predictive mean and model uncertainty) of dropout NNs for selective classification.
\subsection{Mixed-Integer Programming based Selective Classification}
Here, we define a mixed-integer programming model for selective classification to make optimal decisions of classifications and rejections under uncertainty in deep learning. Equivalent to other selective classification models, the aim is to ``reject" making an automated classification for certain instances to increase the performance on non-rejected samples. Similar to many supervised algorithms, our MIP model has two main workflows: training and inference. In the training phase, given an already trained dropout neural network (DNN), we learn the optimal criteria to reject samples by maximizing the accuracy for the non-rejected samples. A natural objective for this task is: 

\begin{center}
\begin{math}
\underset{\phi_{D}, \phi_{R}}{\text{maximize}}
\frac{\sum\limits_{i\in \phi_{D}}{[f(x_{i}) = y_{i}]}}{\sum\limits_{i\in \phi_{D}}{1}}
\end{math}
\end{center}

where $x_{i} \in {\rm I\!R}^{n}$ is the set of features for an instance $i$,\; $y_{i} \in \{0,1\}$ is the label for that instance, and $f : {\rm I\!R}^{n} \rightarrow \{0,1\}$ is the previously trained deep neural network, $\phi_{R}$ is the set of rejected instances, and $\phi_{D}$ is the set of non-rejected (classification decision made) instances.

However, as intuitive this objective function is; it is also non-convex. So, we design a convex proxy by minimizing the number of mistakes made after rejections and utilizing a heuristic regularizer which prevents rejecting samples without increasing the accuracy in the non-rejected sample space. These properties give rise to our objective function as follows: 

\begin{center}
\begin{math}
\underset{\phi_{D}, \phi_{R}}{\text{minimize}}
 \sum\limits_{i\in \phi_{D}}{[f(x_{i}) \neq y_{i}]}
+ \frac{1-\rho}{\sum\limits_{i\in \phi}{1}}\sum\limits_{i\in \phi_{R}}{1}
\end{math}
\end{center}
where $\rho$ is the classification accuracy of the trained deep neural network and $\phi = \phi_{D} \bigcup \phi_{R}$ is the set of all instances. This convex minimization objective exploits the idea that the number of rejections should not exceed the expected number of mistakes if the deep neural network was used without rejection to classify the instances. 

So, what does our model use to determine the rejection population ($\phi_{R}$) and the decision population ($\phi_{D}$)? As introduced by Gal and Ghahramani \cite{gal2016dropout}, our model uses the ``model uncertainty" concept and enhances it with predictive mean to express when the DNN is not confident with its prediction. 

\begin{figure}[h]
    \centering
    \includegraphics[width=\columnwidth]{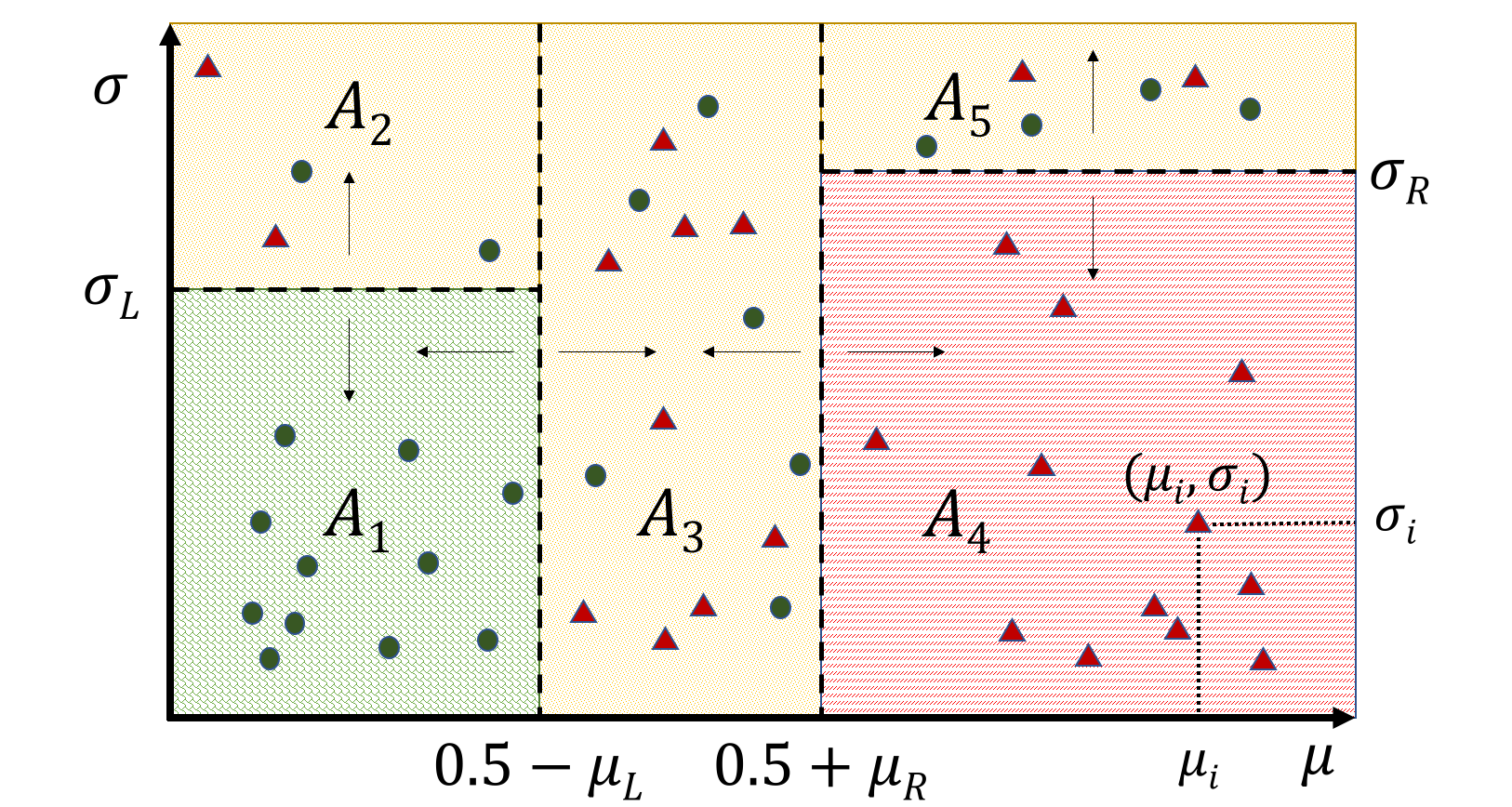}
    \caption{Graphical Illustration of the MIPSC model}
    \label{fig:model2d} 
\end{figure}

For every instance in the training set, we characterize the posterior probability distribution by taking $T$ Monte-Carlo (MC) samples using dropouts. Then, we summarize this predictive distribution by calculating the first moment, predictive mean ($\mu_{i}$) and the second moment, model uncertainty ($\sigma_{i}$) as described by Gal and Ghahramani \cite{gal2016dropout}. Finally, we map the points ($\mu_{i}$,$\sigma_{i}$) for every instance i to a 2D space to characterize the posterior distribution of the whole sample space. One intuitively expects more homogeneous regions to be near lower values of the model uncertainty and extremes of the predictive mean. This intuition can also be observed in Figure \ref{fig:model2d}. Hence, our formulation aims to exploit and optimize upon this structure and identifies the thresholds that define our model's classification and rejection regions. Before formally defining our model, we introduce the notation that we refer to throughout this section in Table \ref{tab:notation1}.

\begin{table}[h]
    \centering
    \begin{tabular}{l|l}
    \toprule
        Var & Definition \\\midrule
        $y_{i}$ & Ground truth label of instance $i$\\
        $p_{i}$  & Positive classification indicator for instance $i$\\
        $n_{i}$  & Negative classification indicator for instance $i$\\
        $r_{i}$  & Rejection indicator for instance $i$\\
        $\mu_{i}$  & Predictive mean for instance $i$\\
        $\sigma_{i}$  & Uncertainty for instance $i$\\
        $\mu_{L}$  & Left boundary for rejection\\
        $\mu_{R}$  & Right boundary for rejection\\
        $\sigma_{L}$  & Upper uncertainty boundary for positive decisions \\
        $\sigma_{R}$  & Upper uncertainty boundary for negative decisions\\
        $L_{i}$ & Left area indicator for instance $i$\\
        $R_{i}$ & Right area indicator for instance $i$\\
        $D_{L_{i}}$ & Down-left area indicator for instance $i$ \\
        $D_{R_{i}}$ & Down-right area indicator for instance $i$\\
        $rCap$ & Rejection capacity \\
        $\rho$ & Deep neural network accuracy without rejection \\
        $M$ & Very large Big-M constant \\
        $\epsilon$ & Very small constant \\
    \end{tabular}
    \caption{Notation Table for MIPSC and MIPCSC}
    \label{tab:notation1}
\end{table}

We indicate the positive classification decision for instance i with $p_{i} \in \{0,1\}$. The negative classification decision for instance i is coded with the variable $n_{i} \in \{0,1\}$. Finally, rejection decision for instance i is characterized with $r_{i} \in \{0,1\}$. For instance i, only one of the indicator variables $p_{i}, n{i}, r_{i}$ can be equal to 1 indicating the decision made for that instance. These decisions are made based on decision thresholds determined by solving the optimization problem defined below. 

We characterize five decision areas of classification and rejection and graphically demonstrate these areas in Figure \ref{fig:model2d}. $A_{1}$ defines the decision region for positive classification, while $A_{4}$ represents the decision region for negative classification. $A_{2}$ and $A_{5}$ are rejection regions due to their high model uncertainty. Thresholds to determine these regions are not tied together for the purpose of handling imbalance or class specific patterns in the data. Finally, $A_{3}$ is another rejection region housing instances having predictive means close to 0.5. In this region, model uncertainty becomes trivial due to its context: it does not matter how "certain" the model is when making a decision similar to a coin toss.  

Boundaries for these regions ($\sigma_{L},\sigma_{R},\mu_{L},\mu_{R}$) are determined by the following set of constraints operating in a supervised fashion through the objective. This is the essential process executed by solving our MIP formulation. 

Here, we start describing our constraints formally. The following constraint regulates the samples which do not reside in the rejection region $A_{3}$ based on their predictive means but on the right hand side of $A_{3}$ such that $i \in A_{4} \cup A_{5}$:
\begin{center}
\begin{equation}
\label{const:Ri}
\mu_{i} > 0.5 + \mu_{R} \text{\;\;iff\;\;} R_{i} = 1
\end{equation}
\end{center}
Now, we would like to distinguish the instances between $A_{4}$ and $A_{5}$ optimally such that our model would make a negative classification decision only when DNN is certain enough. The following constraints characterize the samples that conform to $A_{4}$ such that $i \in A_{4}$:
\begin{center}
\begin{equation}
\label{const:DRi}
\sigma_{i} < \sigma_{R} \text{\;\;iff\;\;} D_{R_{i}} = 1
\end{equation}
\end{center}
\begin{center}
\begin{equation}
R_{i} + D_{R_{i}} > 1 \text{\;\;iff\;\;} n_{i} = 1
\end{equation}
\end{center}
Similarly, the following constraint defines the samples which do not reside in the rejection region $A_{3}$ based on their predictive means but on the left hand side of $A_{3}$ such that $i \in A_{1} \cup A_{2}$:
\begin{center}
\begin{equation}
\label{const:Li}
\mu_{i} < 0.5 - \mu_{L} \text{\;\;iff\;\;} L_{i} = 1
\end{equation}
\end{center}

Further, we would like to distinguish the instances between $A_{1}$ and $A_{2}$ optimally such that our model would make a positive classification decision only when DNN is certain enough. The following constraints characterize the samples that conform to $A_{1}$ such that $i \in A_{1}$:
\begin{center}
\begin{equation}
\label{const:DLi}
\sigma_{i} < \sigma_{L} \text{\;\;iff\;\;} D_{L_{i}} = 1
\end{equation}
\end{center}
\begin{center}
\begin{equation}
L_{i} + D_{L_{i}} > 1 \text{\;\;iff\;\;} p_{i} = 1
\end{equation}
\end{center}

As we have constrained our positive and negative classification decision regions, we reject the remaining instances covered by the constraint below: 
\begin{center}
\begin{equation}
p_{i} + n_{i} + r_{i} = 1
\end{equation}
\end{center}
where the reject decision is assigned when our model cannot make a positive or negative classification decision for instance $i$ due to DNN uncertainty or predictive mean. 

Finally we would like to limit the number of rejections based on the application needs. This is given as: 
\begin{center}
\begin{equation}
\label{const:rCap}
\big(\sum_{i=1}^{m}{r_{i}}\big) \leq rCap
\end{equation}
\end{center}

Combining our objective function and constraints together, then, setting $M$ to be a very large positive constant and fixing $\epsilon$ to be a very small positive constant give rise to the formal definition of our model as follows: 
\begin{align}
\underset{\mu_{L}, \mu_{R}, \sigma_{L}, \sigma_{R}}{\text{minimize}} \sum_{i=1}^{m}{ (p_{i}y_{i} + n_{i}(1-y_{i}))}
+ \frac{1-\rho}{m}\sum_{i=1}^{m}{r_{i}} \;\; \text{s.t.}& \\
\mu_{R} + MR_{i} \geq \mu_{i} - 0.5 \geq \mu_{R} - M(1-R_{i}) + \epsilon, \forall i\\
  M(1-L_{i}) - \mu_{L} - \epsilon \geq \mu_{i} - 0.5 \geq - \mu_{L}  - ML_{i}, \forall i\\
  \sigma_{L} + M(1 - D_{L_{i}}) - \epsilon \geq \sigma_{i} \geq \sigma_{L} - MD_{L_{i}}, \forall i\\
\sigma_{R} + M(1 - D_{R_{i}}) - \epsilon \geq \sigma_{i} \geq \sigma_{R} - MD_{R_{i}}, \forall i\\
D_{L_{i}} + L_{i} \geq 2p_{i} \geq D_{L_{i}} + L_{i} - 1, \forall i \\
D_{R_{i}} + R_{i} \geq 2n_{i} \geq D_{R_{i}} + R_{i} - 1, \forall i \\
p_{i} + n_{i} + r_{i} = 1, \forall i\\
\big(\sum_{i=1}^{m}{r_{i}}\big) \leq rCap \\
\forall p_{i}, r_{i}, n_{i},R_{i},L_{i},D_{L_{i}},D_{R_{i}} \in \{0,1\}\\
\forall  i \in \{1...m\}, and\;\; \mu_{L}, \mu_{R}, \sigma_{L}, \sigma_{R}, \rho \in {\rm I\!R}
\end{align}

In this formulation, constraint (10) is derived from (1), (11) is derived from (4), (12) is derived from (5), (13) is derived from (2), (14) is derived from (6), and (15) is derived from (3) following the Big-M method \cite{griva2009linear}. 

Following the training, the inference is rather straightforward. After acquiring the predictive mean and model uncertainty from DNN for the new sample, a user of our model can arithmetically decide the region the new sample belongs to and make the decision based on the optimal thresholds identified.

\subsection{Cost-Sensitive Selective Classification}

Many classification with reject option problems are cost-sensitive by nature\cite{domingos1999metacost}. An optimal decision-maker should take into various costs and benefits regarding making a correct, an incorrect, or a rejection decision for each case individually.   

For instance, in medical diagnosis, consequences from a false negative decision can be fatal if the diagnosis in question is cancer but not as critical if it is common cold. Within the same context, a doctor can order more tests with varying costs if uncertain depending on the severity of the illness under study.

In fraud management, consequences that arise from not being able to identify a \$5 fraudulent transaction would not be as critical as compared to a \$1000 fraudulent transaction. Moreover, a rejection decision (e.g. initiating a thorough investigation) could be costly, so, might not make financial sense for a \$5 transaction whereas it could be financially sensible for a \$1000 transaction. Finally, the cost of a false positive decision (stopping a good transaction) is not the same as a false negative decision (not catching a fraudulent transaction). All these varying factors need to be considered to come up with an optimal fraud management strategy. 

\begin{figure}[]
    \centering
    \includegraphics[width=\columnwidth]{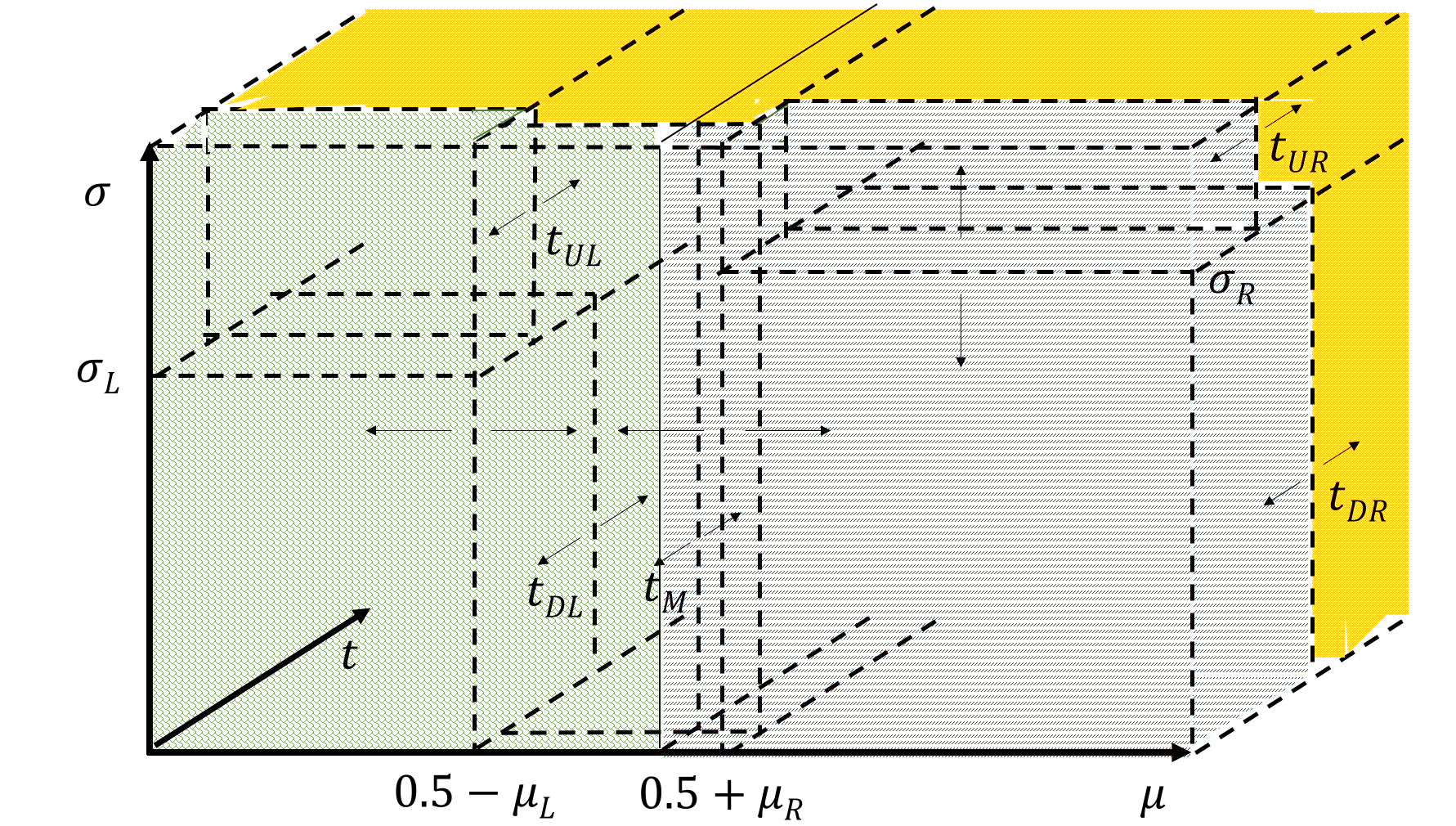}
    \caption{Graphical Illustration of the MIPCSC Model}
    \label{fig:model3d} 
\end{figure}

Thus, we follow Elkan's definition \cite{elkan2001foundations} and extend MIPSC to ``example and class-dependent cost sensitive" settings where each instance belonging to each class has a different cost or benefit of making a correct or incorrect classification and propose Mixed-Integer Programming based Cost-Sensitive Selective Classification (MIPCSC) for cost-sensitive applications. A graphical interpretation of this extension can be viewed in Figure \ref{fig:model3d}. Since the example-dependent value adds another dimension to our problem, we extend the previously introduced five decision regions $(A_{1}, A_{2}, A_{3}, A_{4}, A_{5})$ in Figure \ref{fig:model2d} to three dimensions. Unlike MIPSC, this model would not make a rejection decision just because of the uncertainty in the assessment if it does not make financial sense. Reiterating on the previous example, it is not logical to conduct a fraud investigation that would cost \$50 for a \$5 transaction regardless of the uncertainty. So, in MIPCSC, none of the regions $A_{1}, A_{2}, A_{3}, A_{4}, A_{5}$ are complete rejection regions; but every region $A_{i}$ has a rejection threshold ($t_{DR},t_{UR},t_{DL},t_{UL},t_{M}$) based on the risk associated with it as shown in Figure \ref{fig:model3d}.

Please refer to Table \ref{tab:notation1} and Table \ref{tab:notation2} for the notation used for defining MIPCSC. Formally, we retain our decision variables $\sigma_{L}$,$\sigma_{R}$,$\mu_{L}$,$\mu_{R}$ $\in {\rm I\!R}$ to define the areas ($A_{1}$, $A_{2}$, $A_{3}$, $A_{4}$, $A_{5}$), then introduce five thresholds $t_{DR},t_{UR},t_{DL},t_{UL},t_{M} \in {\rm I\!R}$ for each region based on the third dimension, value (cost/benefit). Instead of having one positive decision as in MIPSC, we define three positive decisions for instance i, $p_{i1}, p_{i2}, p_{i3} \in \{0,1\}$; each corresponding to areas $A_{1}, A_{2}, A_{3}$. Likewise, there are three of negative decisions, $n_{i1}, n_{i2}, n_{i3} \in \{0,1\}$, operating in the regions $A_{5}, A_{4}, and \;\; A_{3}$. Finally, we assign the cost of rejection $c$ to every reject decision.

\begin{table}[h]
    \centering
    \begin{tabular}{c|l}
        Variable & Definition \\\hline
        $\omega_{tp}$ & True-positive decision benefit multiplier\\
        $\omega_{tn}$ & True-negative decision benefit multiplier\\
        $\omega_{fn}$ & False-negative decision cost multiplier\\
        $\omega_{fp}$ & False-positive decision cost multiplier\\
        $t_{i}$ & Value for instance $i$\\
        $c$ & Cost of rejection\\
        $p_{ij}$  & Positive classification indicator for instance $i$ and area $j$\\
        $n_{ij}$  & Negative classification indicator for instance $i$ and area $j$\\
        $t_{DL}$  & Down-left area value boundary for rejection\\
        $t_{UL}$  & Upper-left area value boundary for rejection\\
        $t_{M}$  & Middle area value boundary for rejection\\
        $t_{DR}$  & Down-right area value boundary for rejection\\
        $t_{UR}$  & Upper-right area value boundary for rejection\\
        $S_{DL_{i}}$ & Surface-down-left area indicator for instance $i$ \\
        $S_{DR_{i}}$ & Surface-down-right area indicator for instance $i$\\
        $S_{UL_{i}}$ & Surface-up-left area indicator for instance $i$ \\
        $S_{UR_{i}}$ & Surface-up-right area indicator for instance $i$\\
        $S_{M_{i}}$ & Surface-down-middle area indicator for instance $i$\\
        $Q_{i}$ & Greater than 0.5 indicator for instance $i$
    \end{tabular}
    \caption{Additional Notation Table for MIPCSC}
\label{tab:notation2}
\end{table}

Inheriting constraints (\ref{const:Ri}), (\ref{const:DRi}), (\ref{const:Li}), and (\ref{const:DLi}); we extend our constraints as follows.

The first extra constraint focuses on the region $A_{1}$ and finds the value threshold for rejection in that region. If the DNN output of transaction corresponds to $A_{1}$ region and its example-dependent value is less than the region's value threshold, then our model makes a positive decision. Intuitively, this decision would mean that the expected value of the positive classification decision is superior to the expected value of a rejection decision in $A_{1}$.

\begin{center}
\begin{align}
\label{const:pi1}
t_{i} < t_{DL} \text{\;\;iff\;\;} S_{DL_{i}} = 1 \\
L_{i} + D_{L_{i}} + S_{DL_{i}} > 2 \text{\;\;iff\;\;} p_{i1} = 1
\end{align}
\end{center}

Now, we would like to find our decision threshold for $A_{2}$. Similarly to the previous constraints, if the transaction corresponds to $A_{2}$ region and its value is less than the region's value threshold, then our model makes a positive decision.

\begin{center}
\begin{align}
\label{const:pi2}
t_{i} < t_{UL} \text{\;\;iff\;\;} S_{UL_{i}} = 1 \\
L_{i} + (1 - D_{L_{i}}) + S_{UL_{i}} > 2 \text{\;\;iff\;\;} p_{i2} = 1
\end{align}
\end{center}

Similar to the positive decision regions, now, we focus on the negative decision regions: $A_{4}$ and $A_{5}$. The following constraint focuses in the region $A_{4}$ and finds the value threshold for rejection in that region. If the transaction corresponds to $A_{4}$ region and its value is less than the region's value threshold, then our model makes a negative decision. Intuitively, this decision would mean that the expected value of the negative classification decision is superior to the expected value of a rejection decision in $A_{4}$.

\begin{center}
\begin{align}
\label{const:ni1}
t_{i} < t_{DR} \text{\;\;iff\;\;} S_{DR_{i}} = 1 \\
R_{i} + D_{R_{i}} + S_{DR_{i}} > 2 \text{\;\;iff\;\;} n_{i1} = 1
\end{align}
\end{center}

Now, we would like to find our decision threshold for $A_{5}$. Similarly to the previous constraints, if the transaction corresponds to $A_{5}$ region and its value is less than the region's value threshold, then our model makes a negative decision.

\begin{center}
\begin{align}
\label{const:ni2}
t_{i} < t_{UR} \text{\;\;iff\;\;} S_{UR_{i}} = 1 \\
R_{i} + (1 - D_{R_{i}}) + S_{UR_{i}} > 2 \text{\;\;iff\;\;} n_{i2} = 1
\end{align}
\end{center}

Finally, we move onto our middle region, $A_{3}$. Here, we would like our model to make a positive or negative decision using the predictive mean of 0.5 as the threshold and considering the value threshold we optimally determine by solving the problem, $t_{M}$. 

\begin{center}
\begin{align}
\label{const:pi3ni3}
\mu_{i} > 0.5 \text{\;\;iff\;\;} Q_{i} = 1 \\
t_{i} < t_{M} \text{\;\;iff\;\;} S_{M_{i}} = 1 \\
(2 - L_{i} - R_{i}) + S_{M_{i}} + (1 - Q_{i}) > 3 \text{\;\;iff\;\;} p_{i3} = 1\\
(2 - L_{i} - R_{i}) + S_{M_{i}} + Q_{i} > 3 \text{\;\;iff\;\;} n_{i3} = 1
\end{align}
\end{center}

As we have constrained our positive and negative classification decision regions, we reject the remaining instances covered by the constraint below: 
\begin{center}
\begin{equation}
\sum_{j=1}^{3}{[p_{ij}]} + \sum_{j=1}^{3}{[n_{ij}]} + r_{i} = 1, \forall i\end{equation}
\end{center}
where the reject decision is assigned when making positive or negative classification decision for instance $i$ yields an inferior expected benefit or superior cost compared to a reject decision.

Following our definition and our constraints, we propose our cost-sensitive framework called Mixed-Integer Programming based Cost-Sensitive Selective Classification (MIPCSC) formally as follows:

\begin{align}
\underset{\substack{\mu_{L}, \mu_{R}, \sigma_{L}, \sigma_{R},\\ t_{DL},t_{UL},t_{M},t_{DR},t_{UR}}}{\text{maximize}} &\omega_{tp}(\sum\limits_{i=1}^{n}\sum\limits_{j=1}^{3}{p_{ij}(1-y_{i})t_{i}} + \sum\limits_{i=1}^{n}{r_{i}(1-y_{i})t_{i}}) \nonumber\\
+ & \omega_{tn}(\sum\limits_{i=1}^{m}\sum\limits_{j=1}^{3}{n_{ij}y_{i}t_{i}} + \nonumber \sum\limits_{i=1}^{m}{r_{i}y_{i}t_{i}}) \nonumber\\ - & \omega_{fn}(\sum\limits_{i=1}^{m}\sum\limits_{j=1}^{3}{n_{ij}(1-y_{i})t_{i}}) \nonumber\\ - & \omega_{fp}(\sum\limits_{i=1}^{m}\sum\limits_{j=1}^{3}{p_{ij}y_{i}t_{i}}) -  c\sum\limits_{i=1}^{m}{r_{i}}
\end{align}
\begin{align}
\mu_{R} + MR_{i} \geq \mu_{i} - 0.5 \geq \mu_{R} - M(1-R_{i}) + \epsilon, \forall i\\
  M(1-L_{i}) - \mu_{L} - \epsilon \geq \mu_{i} - 0.5 \geq - \mu_{L}  - ML_{i}, \forall i\\
  \sigma_{L} + M(1 - D_{L_{i}}) - \epsilon \geq \sigma_{i} \geq \sigma_{L} - MD_{L_{i}}, \forall i\\
\sigma_{R} + M(1 - D_{R_{i}}) - \epsilon \geq \sigma_{i} \geq \sigma_{R} - MD_{R_{i}}, \forall i\\
0.5 + MQ_{i} \geq \mu_{i} \geq 0.5 + M(Q_{i}-1) + \epsilon, \forall i\\
t_{DL} + M(1 - S_{DL_{i}}) - \epsilon \geq t_{i} \geq t_{DL} - S_{DL_{i}}, \forall i\\
t_{UL} + M(1 - S_{UL_{i}}) - \epsilon\geq t_{i} \geq t_{UL} - S_{UL_{i}}, \forall i\\
t_{M} + M(1 - S_{M_{i}}) - \epsilon\geq t_{i} \geq t_{M} - S_{M_{i}}, \forall i\\
t_{DR} + M(1 - S_{DR_{i}}) - \epsilon\geq t_{i} \geq t_{DR} - S_{DR_{i}}, \forall i\\
t_{UR} + M(1 - S_{UR_{i}}) - \epsilon\geq t_{i} \geq t_{UR} - S_{UR_{i}}, \forall i \\
D_{L_{i}} + L_{i} + S_{DL_{i}} \geq 3p_{i1}, \forall i \\
D_{L_{i}} + L_{i} + S_{DL_{i}} - 2 \leq 3p_{i1}, \forall i \\
(1 - D_{L_{i}}) + L_{i} + S_{DL_{i}} \geq 3p_{i2}, \forall i \\
    (1 - D_{L_{i}}) + L_{i} + S_{DL_{i}} - 2 \leq 3p_{i2}, \forall i \\
D_{R_{i}} + R_{i} + S_{DR_{i}} \geq 3n_{i1}, \forall i \\
D_{R_{i}} + R_{i} + S_{DR_{i}} - 2 \leq 3n_{i1}, \forall i \\ 
(1 - D_{R_{i}}) + R_{i} + S_{DR_{i}} \geq 3n_{i2}, \forall i \\
(1 - D_{R_{i}}) + R_{i} + S_{DR_{i}} - 2 \leq 3n_{i2}, \forall i\\
(1 - L_{i}) + (1 - R_{i}) + S_{M_{i}} + (1 - Q_{i}) \geq 4p_{i3}, \forall i\\
(1 - L_{i}) + (1 - R_{i}) + S_{M_{i}} + (1 - Q_{i}) - 3 \leq 4p_{i3}, \forall i\\
(1 - L_{i}) + (1 - R_{i}) + S_{M_{i}} + Q_{i} \geq 4n_{i3}, \forall i \\
(1 - L_{i}) + (1 - R_{i}) + S_{M_{i}} + Q_{i} - 3 \leq 4n_{i3}, \forall i \\
\sum_{j=1}^{3}{[p_{ij}]} + \sum_{j=1}^{3}{[n_{ij}]} + r_{i} = 1, \forall i \\
\big(\sum_{i=1}^{m}{r_{i}}\big) \leq rCap \\
\forall p_{ij}, r_{i}, n_{ij},R_{i},L_{i},S_{DL_{i}},DL_{i},DR_{i},S_{UL_{i}},\nonumber\\
S_{M_{i}},S_{DR_{i}},S_{UR_{i}} \in \{0,1\}\\
\mu_{L}, \mu_{R}, \sigma_{L}, \sigma_{R}, t_{DL},  t_{UL},  t_{UL},  t_{UR},  t_{M},\in {\rm I\!R}\\
\forall  i,j \in \{1...m\}
\end{align}

In this formulation, constraint (38) is derived from (28), (39) is derived from (20), (40) is derived from (22), (41) is derived from (29), (42) is derived from (24), and (43) is derived from (26). 

Next, (44) and (45) are derived from (21), (46) and (47) are derived from (23), (48) and (49) are derived from (25), and (50) and (51) are derived from (27).

Finally, (52) and (53) are derived from (30), and (54) and (55) are derived from (31) following the Big-M method, as shown in \cite{griva2009linear}. 

Inference using MIPCSC also follows similar steps to MIPSC. After predictive mean, model uncertainty, and the cost/benefit for the new sample are obtained, a user can arithmetically decide the region the new sample belongs to and make the decision based on the optimal thresholds identified.

\section{EXPERIMENTS}
\subsection{Experimental Setup}
We develop two sets of experiments for classification with reject option and its cost-sensitive extension. For both tasks, we divide the dataset into four distinct sets; the first to train the dropout neural network(DNN), the second to find optimal dropout rate and regularization coefficient to quantify uncertainty, the third to train the proposed MIP models, and the fourth to test the performance of the proposed MIP models. 

To quantify model uncertainty and predictive mean, we train a dropout neural network of 2 hidden layers with relu activations and dropout applied before each layer. For the implementation of DNN, we make use of the source codes of the original authors made publicly available at their website\footnote{https://github.com/yaringal/DropoutUncertaintyExps}. We apply a grid search among dropout rates of (0.05, 0.01, 0.02) and regularization coefficients (0.1, 0.25) to achieve optimal DNN configuration. We run trained and optimized DNN 100 times with its dropouts open. We use empirical mean and standard deviation of these 100 softmax outputs as the predictive mean and the model uncertainty to train our MIP models. We train our MIP models with varying \textit{rCap}'s of [0.1,0.15,0.2,0.25]. We iterate these process 32 times and report average as each model's representative performance. 
\subsection{Evaluation Metrics}
Conventional measures of performance introduced for supervised classification tasks do not represent the performance of a model with reject option under study, comprehensively \cite{reject_metrics}. Here we present four recently introduced metrics for classification with reject option\cite{reject_metrics} and cost sensitive learning\cite{yildirim_aiai}. Ideally, a classifier with reject option should classify as many instances as possible correctly and reject to classify the ones that it would misclassify. A cost-sensitive classifier with reject option makes these decisions based on the profit or loss it would get from each instance. We use $c$ for accurately classified and non-rejected samples, $\overline{c}$ for misclassified and non-rejected samples, $r$ for misclassified and rejected samples, $\overline{r}$ for accurately classified and rejected samples.

\textbf{Non-rejected Accuracy} measures the performance of classification of the model on non-rejected samples. It is defined as;
\begin{equation*}
\dfrac{c}{c+\overline{c}}
\end{equation*}

\textbf{Classification Quality} measures the performance of both classification and rejection of the model. It is defined as 
\begin{equation*}
\dfrac{c+r}{c+r+\overline{c}+\overline{r}}
\end{equation*}

\textbf{Rejection Quality} measures the relative performance of rejection to the overall performance of classification. It is defined as; 
\begin{equation*}
\dfrac{r/\overline{r}}{(\overline{c}+r)/(c+\overline{r})}
\end{equation*}

\textbf{Profit Gain} measures the level of gained profit from the model outcome relative to perfectly classifying every instance without any rejection and assigning every instance to the majority class. Let $\$_{model}$ be the profit gain of the model under study, $\$_{oracle}$ be the profit gain of the perfect model, and $\$_{majority}$ be the profit gain of the majority class assigning model, we define profit gain as;

\begin{equation*}
\dfrac{\$_{model} - \$_{majority}}{\$_{oracle} - \$_{majority}}
\end{equation*}

\subsection{Experiments with UCI Datasets}
In this section, we discuss how the performance of our framework MIPSC is on several publicly available datasets. We experiment with ten datasets from UCI classification repository. We set up experiments of binary classification with reject option on datasets coming from various application areas. We refer readers to Table \ref{tab:uci_dataset} for simple statistics of datasets. They span applications of credit card applications(\textit{australian}), medical diagnosis(\textit{breast, diabetic, heart, haberman, pima}), housing prices (\textit{house}) and discriminating various types of signals (\textit{ionosphere, seismic, sonar}). The variety of imbalance from 53\% to 93\% among our datasets also helps us to stress our framework to label imbalances.
\begin{table}[]
    \centering
    \begin{tabular}{l c c c }
        \toprule
          & Instances & Features & Majority Class\\
          \midrule
         \textit{australian} & 689 & 14 & 67\% \\
         \textit{breast} & 699 & 19 & 65\% \\
         \textit{diabetic} & 1151 & 19 & 54\% \\
         \textit{haberman} & 306 & 4 & 74\% \\
         \textit{heart} & 303 & 20 & 54\% \\
         \textit{house} & 435 & 48 & 61\% \\
         \textit{ionoshpere} & 351 & 34 & 65\% \\
         \textit{pima} & 768 & 8 & 66\% \\
         \textit{seismic} & 2584 & 18 & 93\% \\
         \textit{sonar} & 208 & 60 & 53\% \\
         \bottomrule
    \end{tabular}
    \caption{UCI Dataset Statistics}
    \label{tab:uci_dataset}
\end{table}

\subsubsection{Baselines}
We compare the MIPSC with three other baselines.

\begin{itemize}
\item\textbf{Random} baseline chooses samples to reject randomly.

\item\textbf{Predictive mean} baseline chooses the closest samples to have 0.5 predictive mean to be rejected until the rejection capacity parameter is met\cite{chow1970optimum,svmreject}.

\item\textbf{Model uncertainty} baseline chooses the samples with the highest standard deviation to be rejected until the rejection capacity parameter is met \cite{gal2016dropout,leibig2017leveraging}.
\end{itemize}
Comparison with \textit{random} baseline helps us to investigate if using \textit{predictive mean} or \textit{model uncertainty} adds any value to find optimal decisions when rejecting. Comparing MIPSC with \textit{predictive mean} and \textit{model uncertainty} separately allows us to investigate if they are complementary in optimal decision making for classification with reject option.

\subsubsection{Results}

\begin{figure*}[ht]
\centering
\rule{\textwidth}{1pt}\\
\begin{subfigure}{.41\columnwidth}
\includegraphics[width=\columnwidth]{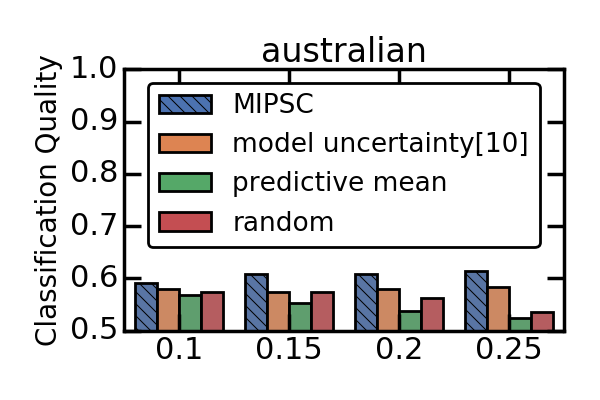}
\end{subfigure}
\begin{subfigure}{.41\columnwidth}
\includegraphics[width=\columnwidth]{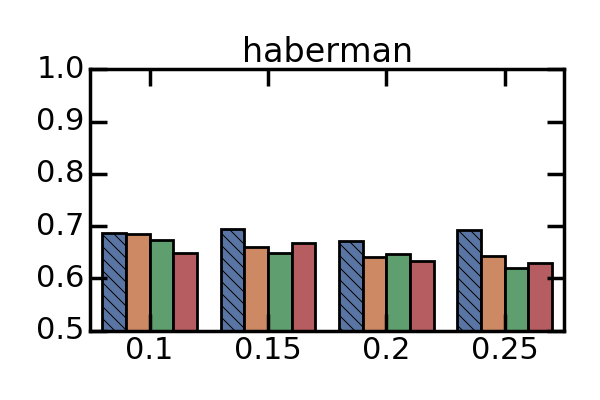}
\end{subfigure}
\begin{subfigure}{.41\columnwidth}
\includegraphics[width=\columnwidth]{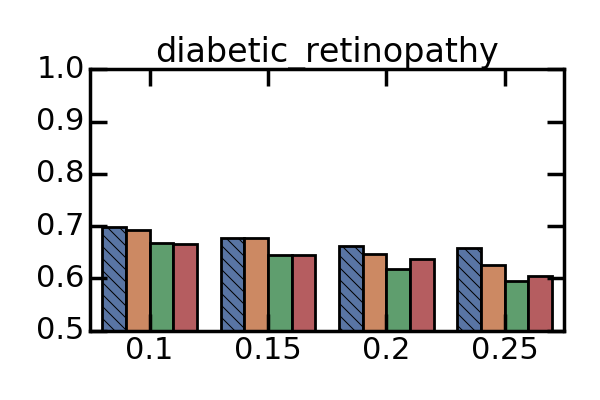}
\end{subfigure}
\begin{subfigure}{.41\columnwidth}
\includegraphics[width=\columnwidth]{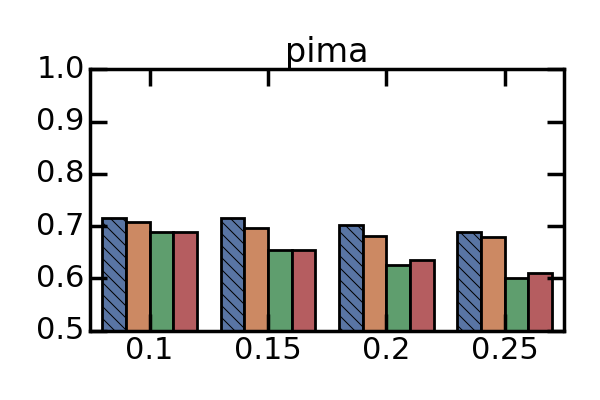}
\end{subfigure}
\begin{subfigure}{.41\columnwidth}
\includegraphics[width=\columnwidth]{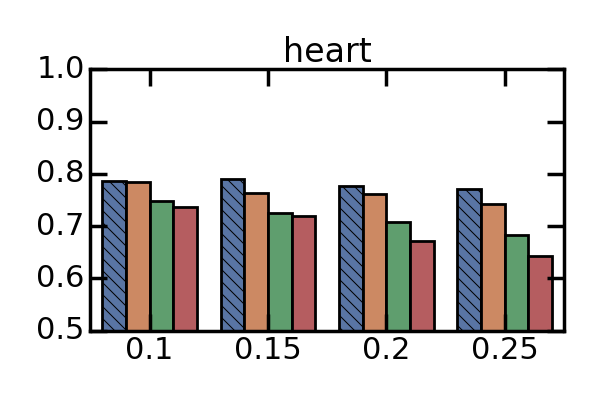}
\end{subfigure}
\begin{subfigure}{.41\columnwidth}
\includegraphics[width=\columnwidth]{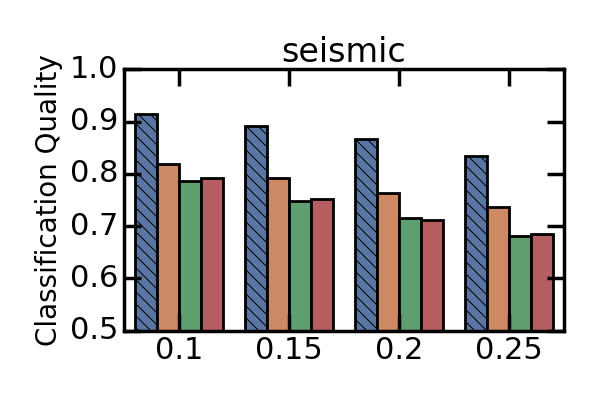}
\end{subfigure}
\begin{subfigure}{.41\columnwidth}
\includegraphics[width=\columnwidth]{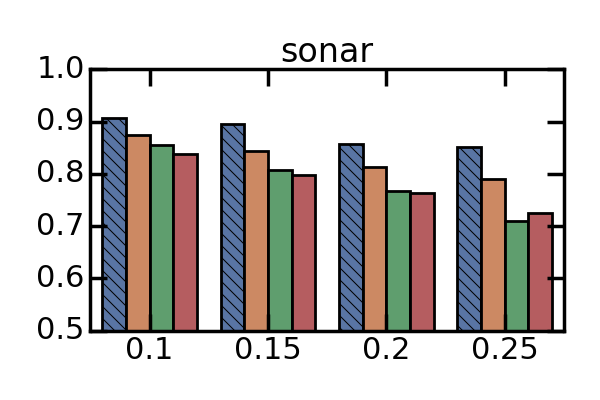}
\end{subfigure}
\begin{subfigure}{.41\columnwidth}
\includegraphics[width=\columnwidth]{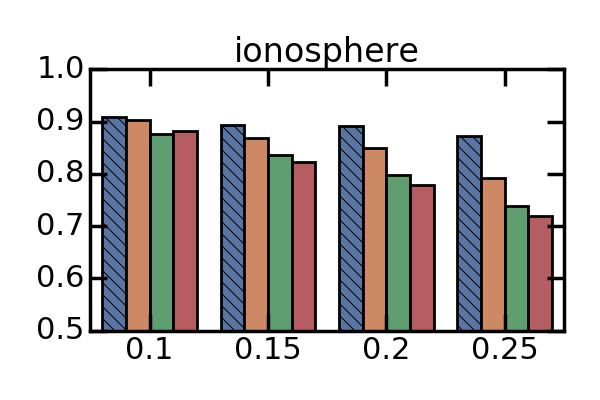}
\end{subfigure}
\begin{subfigure}{.41\columnwidth}
\includegraphics[width=\columnwidth]{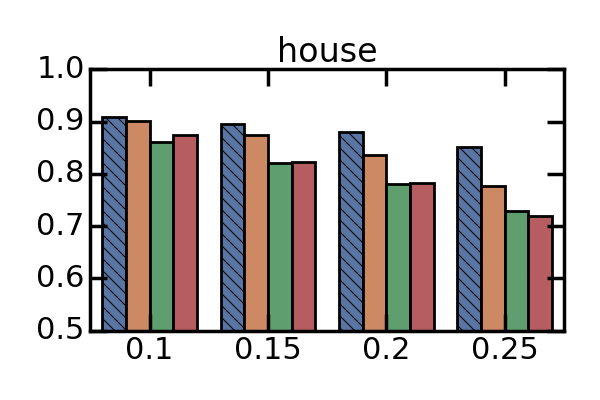}
\end{subfigure}
\begin{subfigure}{.41\columnwidth}
\includegraphics[width=\columnwidth]{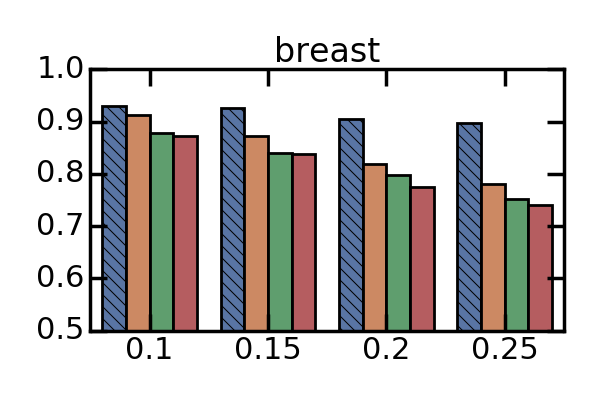}
\end{subfigure}\\
\rule{\textwidth}{1pt}\\
\begin{subfigure}{.41\columnwidth}
\includegraphics[width=\columnwidth]{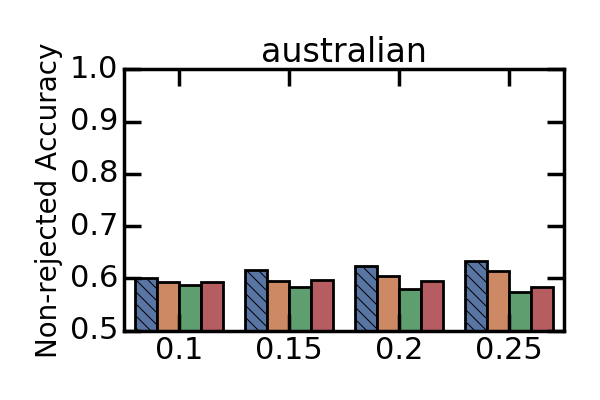}
\end{subfigure}
\begin{subfigure}{.41\columnwidth}
\includegraphics[width=\columnwidth]{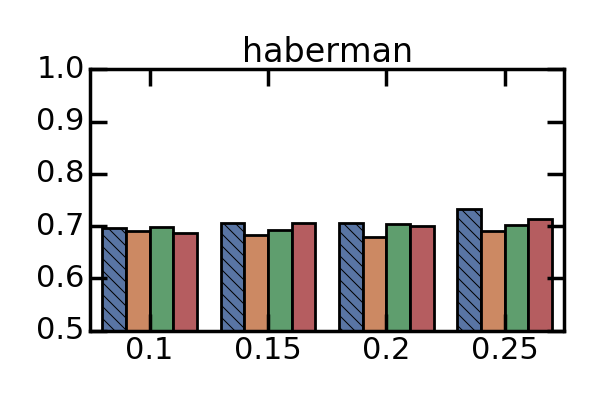}
\end{subfigure}
\begin{subfigure}{.41\columnwidth}
\includegraphics[width=\columnwidth]{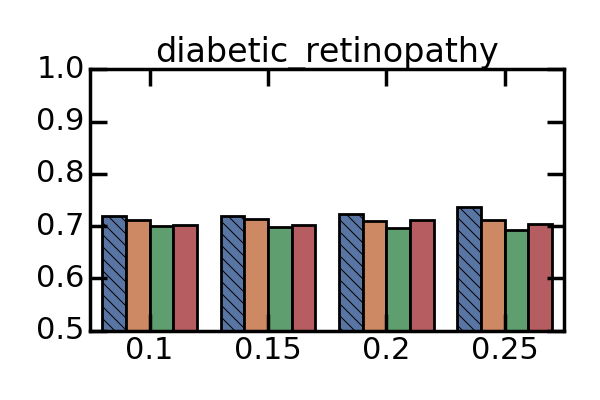}
\end{subfigure}
\begin{subfigure}{.41\columnwidth}
\includegraphics[width=\columnwidth]{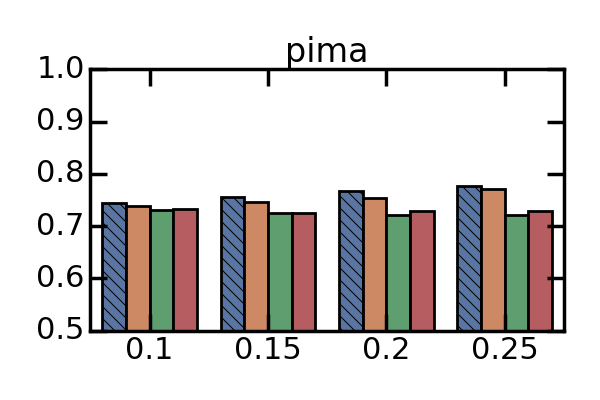}
\end{subfigure}
\begin{subfigure}{.41\columnwidth}
\includegraphics[width=\columnwidth]{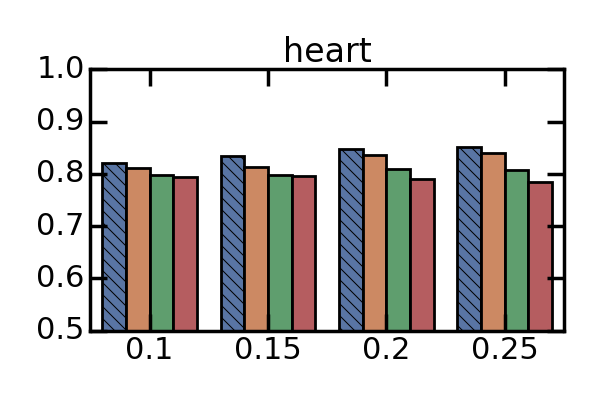}
\end{subfigure}
\begin{subfigure}{.41\columnwidth}
\includegraphics[width=\columnwidth]{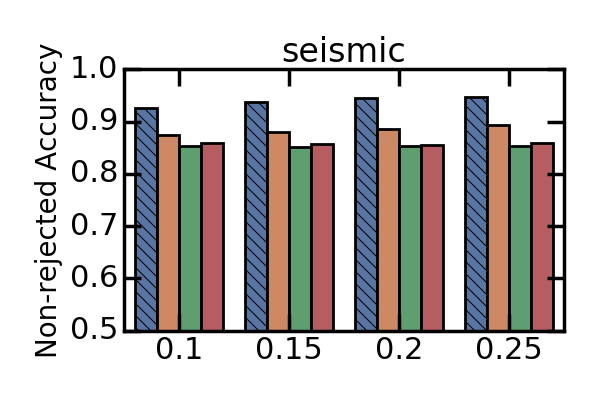}
\end{subfigure}
\begin{subfigure}{.41\columnwidth}
\includegraphics[width=\columnwidth]{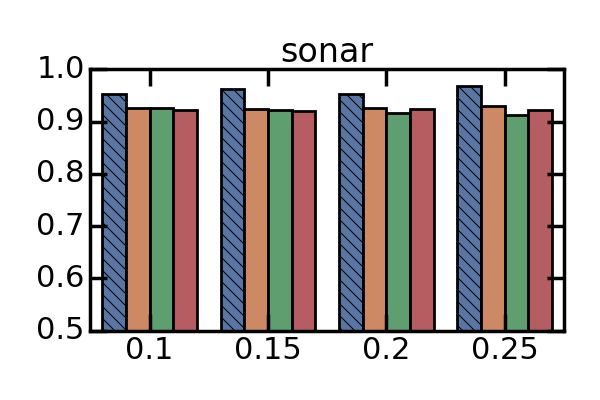}
\end{subfigure}
\begin{subfigure}{.41\columnwidth}
\includegraphics[width=\columnwidth]{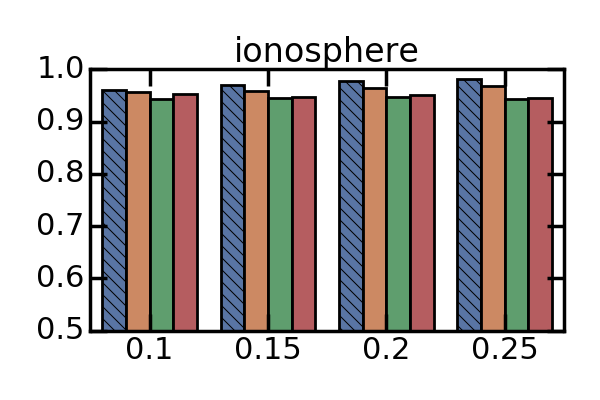}
\end{subfigure}
\begin{subfigure}{.41\columnwidth}
\includegraphics[width=\columnwidth]{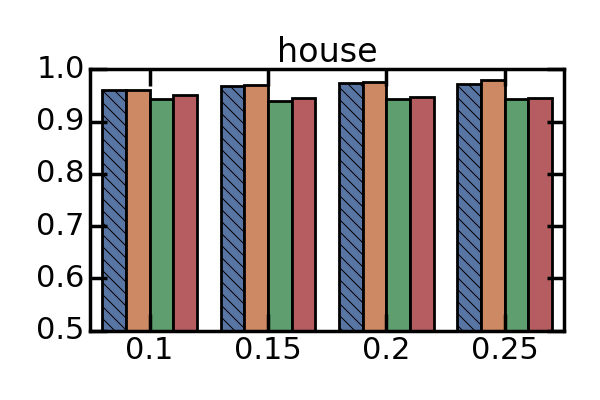}
\end{subfigure}
\begin{subfigure}{.41\columnwidth}
\includegraphics[width=\columnwidth]{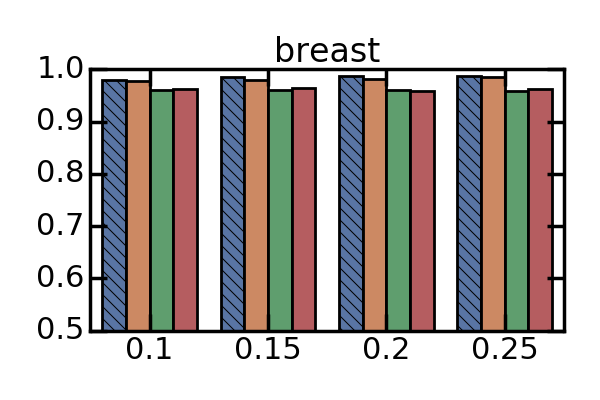}
\end{subfigure}\\
\rule{\textwidth}{1pt}\\
\begin{subfigure}{.41\columnwidth}
\includegraphics[width=\columnwidth]{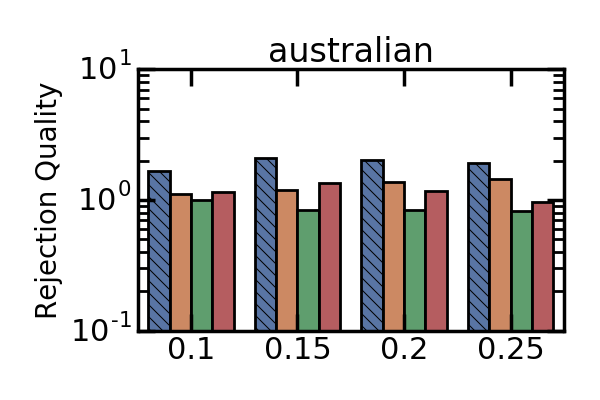}
\end{subfigure}
\begin{subfigure}{.41\columnwidth}
\includegraphics[width=\columnwidth]{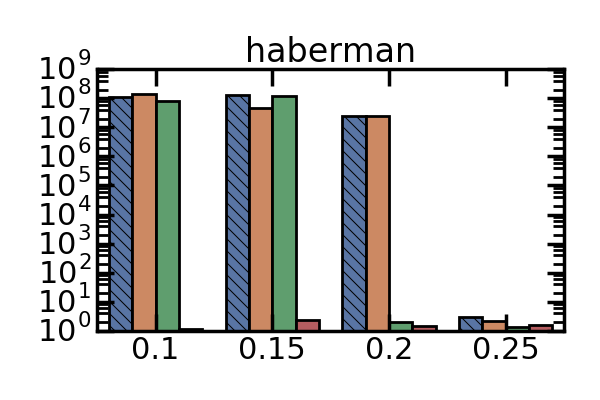}
\end{subfigure}
\begin{subfigure}{.41\columnwidth}
\includegraphics[width=\columnwidth]{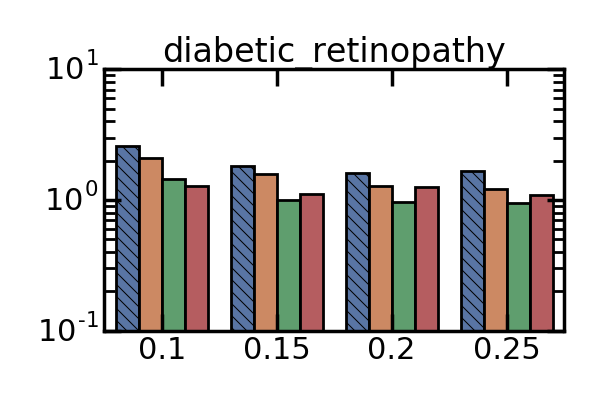}
\end{subfigure}
\begin{subfigure}{.41\columnwidth}
\includegraphics[width=\columnwidth]{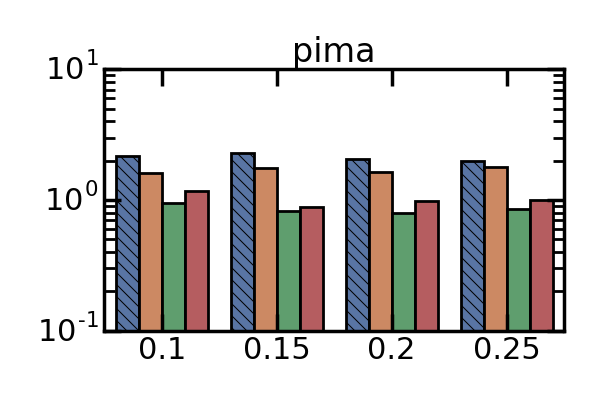}
\end{subfigure}
\begin{subfigure}{.41\columnwidth}
\includegraphics[width=\columnwidth]{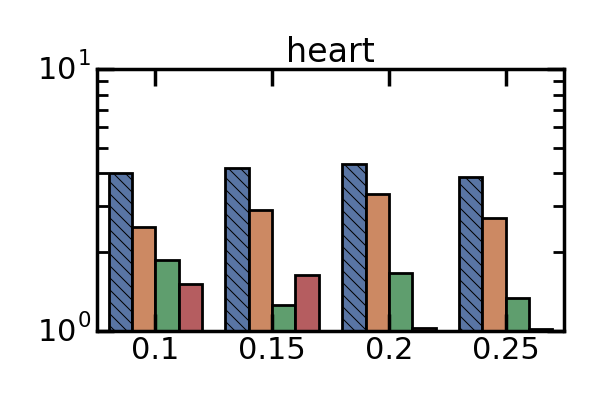}
\end{subfigure}
\begin{subfigure}{.41\columnwidth}
\includegraphics[width=\columnwidth]{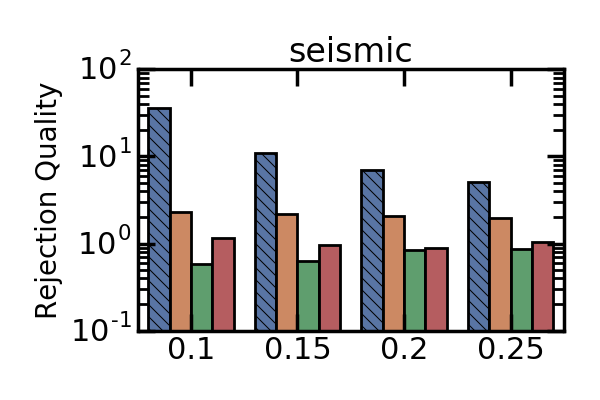}
\end{subfigure}
\begin{subfigure}{.41\columnwidth}
\includegraphics[width=\columnwidth]{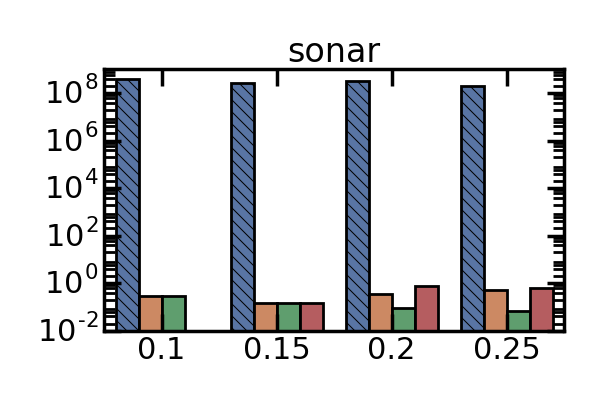}
\end{subfigure}
\begin{subfigure}{.41\columnwidth}
\includegraphics[width=\columnwidth]{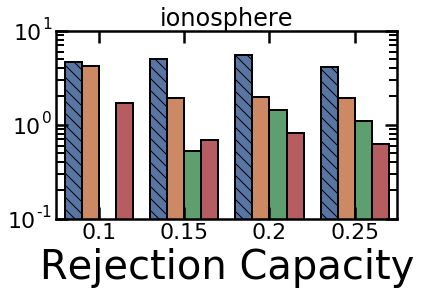}
\end{subfigure}
\begin{subfigure}{.41\columnwidth}
\includegraphics[width=\columnwidth]{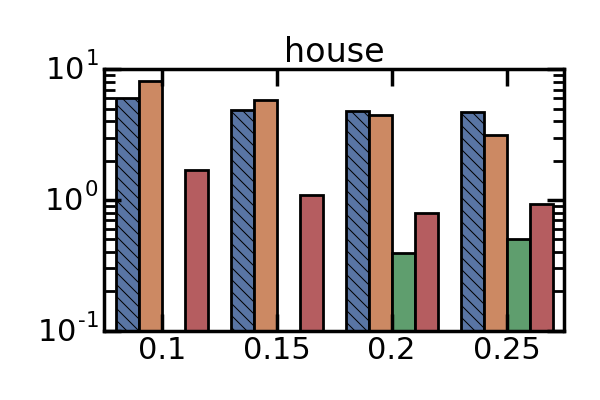}
\end{subfigure}
\begin{subfigure}{.41\columnwidth}
\includegraphics[width=\columnwidth]{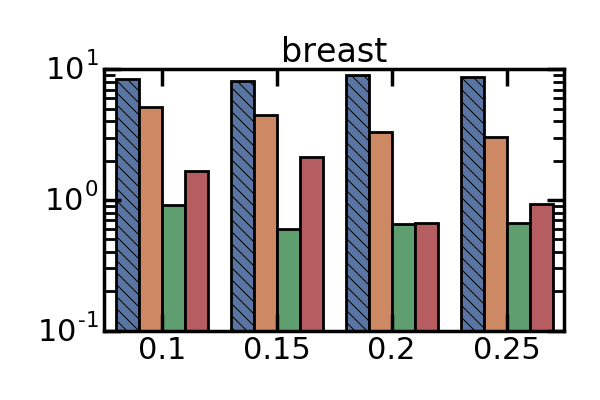}
\end{subfigure}\\
\rule{\textwidth}{1pt}\\
\caption{\textbf{Performance of the MIPSC and other baselines under varying rejection capacities. Notice the superior performance of MIPSC over the recent state-of-the-art and other baselines in all of the three performance metrics for various publicly available datasets.}}
\label{fig:uci_results}
\end{figure*}
Figure \ref{fig:uci_results} shows the performance of MIPSC, and the other baselines. We make the following major observations; 

\begin{itemize}

\item In all ten datasets and three evaluation metrics, MIPSC almost always achieves superior performance compared to the baselines,

\item Higher rejection capacities yield higher non-rejected accuracy, as expected,

\item Model uncertainty baseline consistently performs the second best signaling a better characterization of rejection than predictive mean, 

\item Predictive mean can be complementary in classification with rejection task as evidenced by MIPSC higher performance than model uncertainty baseline. By itself, predictive mean baseline achieves similar performances to the random baseline.

\end{itemize}

As noted earlier, these ten datasets have varying degrees of imbalances. It is promising to see the consistent superiority of our framework compared to the baselines regardless of the imbalances. Our approach achieves up to 15.01\% higher classification quality score, up to 6.65\% higher non-rejected accuracy, and higher rejection quality score at billion scale (note that rejection capacity is an unbounded metric). Even though our method achieves better classification quality score in the house dataset, model uncertainty baseline appears to have higher non-rejected accuracy and rejection quality. We attribute this phenomenon to overfitting of our model (100\% training dataset non-rejected accuracy and classification quality), and it will constitute part of our future work for introducing regularization for our model.

\subsection{Online Fraud Management}
In this section, we discuss the contribution of our cost-sensitive framework MIPCSC over industry-standard baselines in online fraud management tasks. We design our experiments with three real-world e-commerce one week worth of online transaction datasets coming from digital goods, office supplies, and sporting goods stores. Summary statistics of our datasets can be seen in Table \ref{tab:fraud_dataset}.

\begin{table}[]
    \centering
    \begin{tabular}{l c c c}
    \toprule
        Store & Transactions & Fraud-Ratio & Avg. Amount(\$)\\
        \midrule
        Digital Goods & 67,215 & 8.1\% & \$79.29\\
        Office Supplies & 10,678 & 17.2\% & \$330.10\\
        Sporting Goods & 6,968 & 3.5\% & \$296.34\\
        \bottomrule
    \end{tabular}
    \caption{Online Purchase Transactions Dataset Statistics}
    \label{tab:fraud_dataset}
\end{table}

In online fraud management, our base task is to classify each transaction instance as legitimate or fraudulent. Different than a standard classification task, benefits and costs of each true and false classification vary with the transaction amount of each transaction instance. Moreover, true classification of a legitimate transaction and a fraudulent transaction do not bring the same amount of benefit. False classification of a legitimate transaction or a fraudulent transaction incurs different costs as well (i.e., customer insult, fraud loss). 
\subsubsection{Parameter Settings}
Our task also involves rejecting making classification when uncertain. In online fraud management domain, ``rejecting to make a decision" equates to sending the transaction instance to an expert to be reviewed. This process of rejecting to make a decision also comes with a cost. By taking all these aforementioned costs and benefits of the task into consideration, here we present the final profit gain that our framework and several other fraud management strategies achieve on three real-world datasets. For our experiments we set the model parameters $c$ as 3, $\omega_{tp}$ as 0.2, $\omega_{fp}$ as 2.4, and $\omega_{fn}$ as 3 following business standards from well-accredited business reports\footnote{CyberSource: 2017 North America online fraud benchmark Report. Report. CyberSource Corporation (2017)}. 

We note that $\omega_{tp}$ corresponds to classifying legitimate transactions correctly, thus profit gain. $\omega_{fp}$ corresponds to classifying fraudulent transactions as legitimate, thus fraud loss. In our setting, when a merchant is not able to detect fraudulent transaction (false positive), it is liable for the product's cost itself, shipping costs, and extra chargeback fees. $\omega_{fn}$ corresponds to classifying a legitimate transaction as fraudulent, thus consumer insult. Stopping customer's valid transaction requests due to fraudulent activity usually results in losing that customer's future lifetime transactions. We set $\omega_{tn}$ as 0, since preventing fraud does not provide any discernible financial benefit.
\subsubsection{Baselines}
We compare the MIPCSC with four other baselines. We adopt two of them from the previous section (model uncertainty and random) and introduce two new cost-sensitive baselines.

\begin{itemize}

\item\textbf{Transaction amount} baseline rejects to classify the instances with the largest transaction amounts until the rejection capacity is met. Majority of the transaction processors follows this conservative strategy.

\item\textbf{Risk} baseline rejects to classify the instances based on both model uncertainty and transaction amount. It multiplies the model uncertainty and transaction amount and rejects to classify the ones with the highest value until the rejection capacity is met.

\end{itemize}

Comparing MIPCSC with transaction amount baseline helps to assess whether our approach performs better than the most conservative fraud management strategy. Comparing MIPCSC with risk baseline assist with understanding if our approach is capable of making better assessments of cost-sensitive decisions than a simple arithmetic cost-sensitive risk measurement.

\begin{figure}
\begin{subfigure}{.8\columnwidth}
\includegraphics[width=\columnwidth]{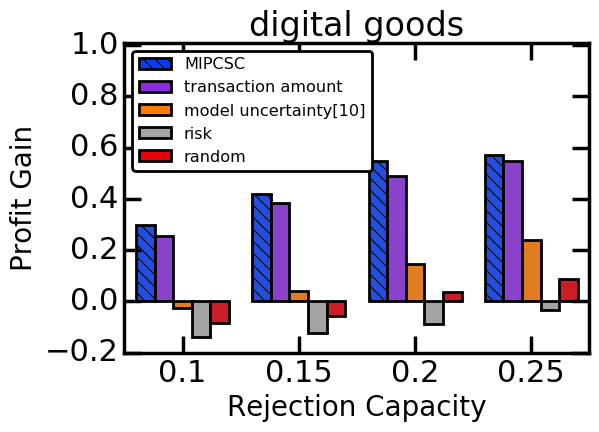}
\end{subfigure}
\begin{subfigure}{.49\columnwidth}
\includegraphics[width=\columnwidth]{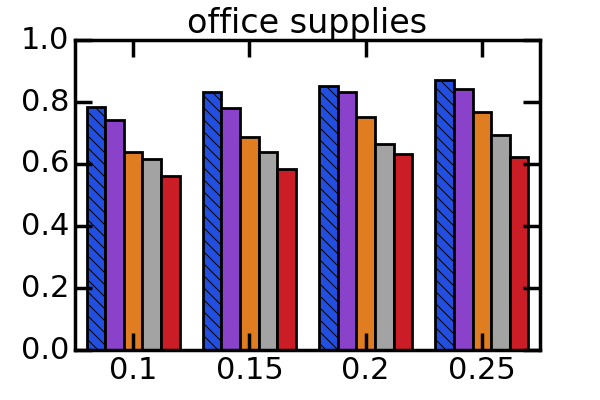}
\end{subfigure}
\begin{subfigure}{.49\columnwidth}
\includegraphics[width=\columnwidth]{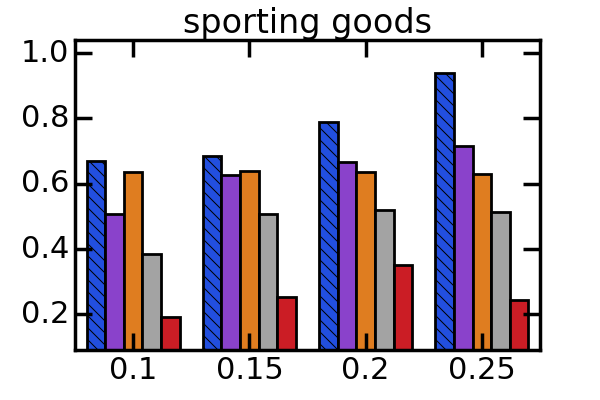}
\end{subfigure}
\centering
\caption{\textbf{Profit Gain of MIPCSC vs. baselines for fraud management.}}
\label{fig:profit_results}
\end{figure}

\subsubsection{Results}
Figure \ref{fig:profit_results} shows the performance of MIPCSC compared to the other baselines. Our key observations are given as follows: 

\begin{itemize}

\item Under varying capacities of rejection, MIPCSC always achieves the highest profit gain, 

\item In the Digital Goods dataset, underlying DNN performs worse than outputting a trivial solution, thus causes uncertainty based baselines to obtain negative profit gain at various rejection capacities. It is clear that MIPCSC is robust to the underlying DNN performance giving the highest profit gain in all cases, 

\item Constant inferior performance of the Risk baseline suggests that simply combining uncertainty with a value aspect does not help to make a cost-optimal decision. 
\end{itemize}

The necessity of a framework like MIPCSC becomes apparent observing its constant effectiveness. It achieves up to 23\% higher profit gain compared to the business standards of fraud management today. This translates to saving up to \$40,570 in the retailer's weekly online business. Extrapolating the amount to yearly revenue gain (assuming uniform distribution over weeks), MIPCSC provides extra \$2M profit gain compared to its closest competitor baseline.

\section{CONCLUSION}

In this paper, we introduced MIPSC: a novel and extensible selective classification model that effectively utilizes uncertainty in deep learning and combines it with predictive mean to make optimal decisions. We demonstrated MIPSC's effectiveness using state-of-the-art selective classification metrics in publicly available datasets from various domains. We found that predictive mean is complementary to model uncertainty for making optimal reject decisions. Furthermore, we showcased a real-world use-case of online fraud management using our cost-sensitive extension, MIPCSC. Future work includes (1) experimenting with other Bayesian frameworks, (2) optimizing the MIP performance by designing novel column generation techniques, and (3) extending the frameworks to non-binary settings. 

\bibliographystyle{ACM-Reference-Format}
\bibliography{ijcai19}


\end{document}